\documentclass[letterpaper]{article}
\usepackage{proceed2e}
\usepackage[margin=1in]{geometry}

\usepackage{tabularx}
\usepackage{times}
\usepackage{mathtools, cuted}
\usepackage{subfigure} 
\usepackage{booktabs}       
\usepackage{amsfonts}       
\usepackage{nicefrac}       
\usepackage{microtype}      
\usepackage{amsmath,amssymb,amsopn}
\usepackage{natbib}

\newcommand{\N}{\mathcal{N}}
\newcommand{\bM}{\mathcal{M}}
\newcommand{\Log}{\mathcal{L}}

\newcommand{\E}{\mathbb{E}}
\newcommand{\x}{{\mathbf{x}}}

\newcommand{\btheta}{{\boldsymbol\theta}}

\newcommand{\bZ}{{\mathbf{Z}}}
\newcommand{\bSigma}{{\boldsymbol\Sigma}}

\newcommand{\X}{\mathbf {X}}
\newcommand{\bB}{\mathbf {B}}
\newcommand{\bG}{\mathbf {G}}

\newcommand{\bb}{\mathbf {b}}
\newcommand{\bd}{\mathbf {d}}
\newcommand{\bv}{\mathbf {v}}
\newcommand{\bw}{\mathbf {w}}

\newcommand{\bV}{\mathbf {V}}
\newcommand{\bD}{\mathbf {D}}
\newcommand{\bW}{\mathbf {W}}

\newcommand{\brwd}{{\boldsymbol r}}

\newcommand{\bfv}{{\boldsymbol f}}

\newcommand{\R}{\mathbb{R}}

\newcommand{\bX}{{\mathbf{X}}}
\newcommand{\bI}{\mathbf{I}}
\newcommand{\bzero}{\mathbf{0}}

\newcommand{\bftheta}{{\boldsymbol \theta}}

\newcommand{\ie}{i.e.,\ }
\newcommand{\eg}{e.g.,\ }

\usepackage{cleveref}
\crefname{section}{§}{§§}
\Crefname{section}{§}{§§}
\setcitestyle{authoryear,round,citesep={;},aysep={,},yysep={;}}

\title{Inverse Reinforcement Learning via Deep Gaussian Process}

\author{{Ming Jin}\thanks{This research is funded by the Republic of Singapore's National Research Foundation through a grant to the Berkeley Education Alliance for Research in Singapore (BEARS) for the Singapore-Berkeley Building Efficiency and Sustainability in the Tropics (SinBerBEST) Program. BEARS has been established by the University of California, Berkeley as a center for intellectual excellence in research and education in Singapore.}\\
EECS, UC Berkeley, USA\\
{jinming@berkeley.edu}
\And
{Andreas Damianou}\thanks{Work done while this author was at the University of Sheffield.}\\
Amazon.com, Cambridge, UK\\
damianou@amazon.com
\And
{Pieter Abbeel}\\
EECS, UC Berkeley, USA\\
pabbeel@berkeley.edu 
\And
{Costas Spanos}\\
EECS, UC Berkeley, USA\\
spanos@berkeley.edu
}
\begin{document}

\maketitle

\begin{abstract}
We propose a new approach to inverse reinforcement learning (IRL) based on the deep Gaussian process (deep GP) model, which is capable of learning complicated reward structures with few demonstrations. Our model stacks multiple latent GP layers to learn abstract representations of the state feature space, which is linked to the demonstrations through the Maximum Entropy learning framework. Incorporating the IRL engine into the nonlinear latent   structure renders existing deep GP inference approaches intractable. To tackle this, we develop a non-standard variational approximation framework which extends previous inference schemes. This allows for approximate Bayesian treatment of the feature space and guards against overfitting. Carrying out representation and inverse reinforcement learning simultaneously within our model outperforms state-of-the-art approaches, as we demonstrate with experiments on standard benchmarks (``object world'',``highway driving'') and a new benchmark (``binary world'').
\end{abstract}
\section{INTRODUCTION}
\label{sec:intro}

The problem of inverse reinforcement learning (IRL) is to infer the latent reward function that the agent subsumes by observing its demonstrations or trajectories in the task. It has been successfully applied in scientific inquiries, \eg animal and human behavior modeling \citep{ng2000algorithms}, as well as practical challenges, \eg navigation \citep{ratliff2006maximum,abbeel2008apprenticeship,ziebart2008maximum} and intelligent building controls \citep{barrett2015autonomous}. By learning the reward function, which provides the most succinct and transferable definition of a task, IRL has enabled advancing the state of the art in the robotic domains \citep{abbeel2004apprenticeship,kolter2007hierarchical}.

Previous IRL algorithms treat the underlying reward as a linear \citep{abbeel2004apprenticeship,ratliff2006maximum,ziebart2008maximum,syed2007game, ratliff2009learning} or non-parametric function \citep{levine2010feature,levine2011nonlinear} of the state features. Main formulations within the linearity category include maximum margin \citep{ratliff2006maximum}, which presupposes that the optimal reward function leads to maximal difference of expected reward between the demonstrated and random strategies, and feature expectation matching \citep{abbeel2004apprenticeship,syed2008apprenticeship}, based on the observation that it suffices to match the feature expectation of a policy to the expert in order to guarantee similar performances. The reward function can be also regarded as the parameters for the policy class, such that the likelihood of observing the demonstrations is maximized with the true reward function, \eg the maximum entropy approach \citep{ziebart2008maximum}. 

As the representation power is limited by the linearity assumption, nonlinear formulations \citep{levine2010feature} are proposed to learn a set of composite features based on logical conjunctions. Non-parametric methods, pioneered by \citep{levine2011nonlinear} based on Gaussian Processes (GPs) \citep{rasmussen2006gaussian}, greatly enlarge the function space of latent reward to allow for non-linearity, and have been shown to achieve the state of the art performance on benchmark tests, \eg object world and simulated highway driving \citep{abbeel2004apprenticeship,syed2007game,levine2010feature,levine2011nonlinear}. Nevertheless, the heavy reliance on predefined or handcrafted features becomes a bottleneck for the existing methods, especially when the complexity or essence of the reward can not be captured by the given features.
 Finding such features automatically from data would be highly desirable.

In this paper, we propose an approach which performs feature and inverse reinforcement learning simultaneously and coherently within the same model by incorporating deep learning. The success of deep learning in a wide range of domains has drawn the community's attention to its structural advantages that can improve learning in complicated scenarios, \eg \cite{mnih2013playing} recently achieved a deep reinforcement learning (RL) breakthrough. Nevertheless, most deep models require massive data to be properly trained and can become impractical for IRL. 
On the other hand, deep Gaussian processes (deep GPs) \citep{damianou2013deep,damianou2015deep} not only can they learn abstract structures with \emph{smaller} data sets, but they also retain the non-parametric properties which \cite{levine2011nonlinear} demonstrated as important for IRL. 

A deep GP is a deep belief network comprising a hierarchy of latent variables with Gaussian process mappings between the layers. 
Analogously to how gradients are propagated through a standard neural network, deep GPs aim at propagating uncertainty through Bayesian learning of latent posteriors. This constitutes a useful property for approaches involving stochastic decision making and also guards against overfitting by allowing for noisy features. However, previous methodologies employed for approximate Bayesian learning of deep GPs \citep{damianou2013deep,hensman2014nested,bui2015training,mattos:RGP16} fail when diverging from the simple case of fixed  output data modeled through a Gaussian regression model. In particular, in the IRL setting, the reward (output) is only revealed through the demonstrations, which is guided by the policy given by the reinforcement learning \citep{damianou2015deep}.

\begin{figure*}[t]
\vskip 0.2in
\centering
\begin{minipage}{.48\textwidth}
  \centering
  {\includegraphics[width=\columnwidth, trim = 62mm 120mm 30mm 30mm, clip]{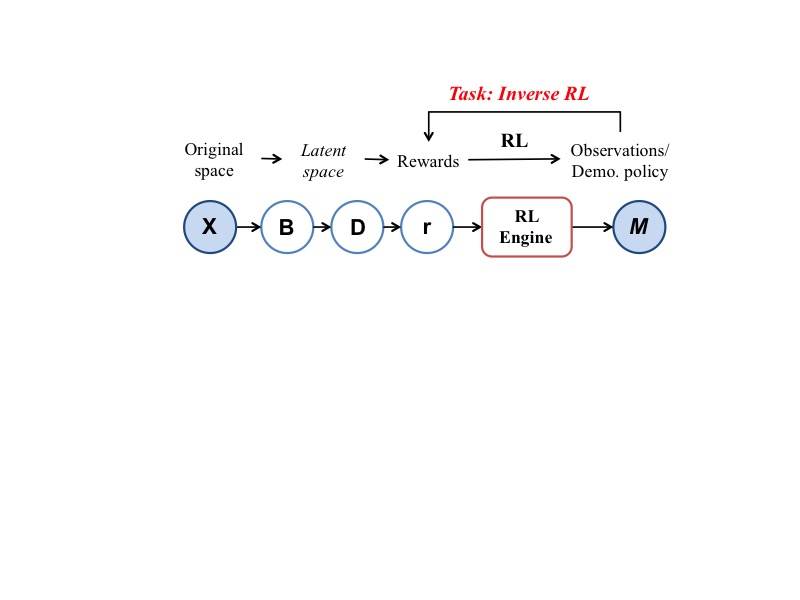}}
  \label{fig:gpirl_overview}
\end{minipage}
\hfill
\begin{minipage}{.48\textwidth}
  \centering
 {\includegraphics[width=\columnwidth, trim = 50mm 120mm 45mm 30mm, clip]{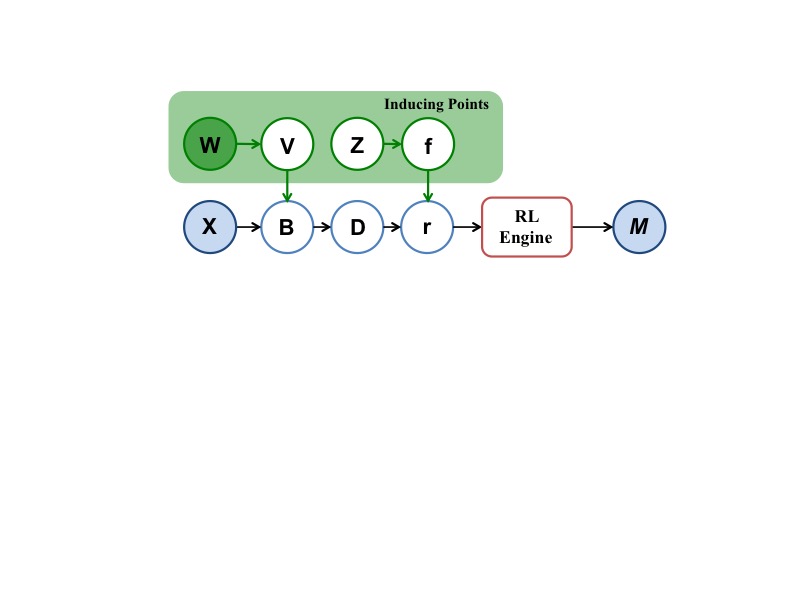}}
\label{fig:gpirl_new}
\end{minipage}
\caption{\textit{Left: }The proposed deep GP model for IRL. The two layers of GPs are stacked to generate the latent reward $\brwd$, which is input to the reinforcement learning (RL) engine to produce an optimal control policy and demonstrations. $\X$ denotes the initial feature representation of states. \textit{Right: }Illustration of DGP-IRL augmented with inducing outputs $\bfv,\bV$ and corresponding inputs $\bZ,\bW$.}
\end{figure*}

The main contributions of our paper are below.
\begin{itemize}
\item We extend the deep GP framework to the IRL domain (Fig. \ref{fig:gpirl_overview}), allowing for learning latent rewards with more complex structures from limited data.
\item We derive a variational lower bound on the marginal log likelihood using an innovative definition of the variational distributions. This methodological contribution enables Bayesian learning in our model and can be applied to other scenarios where the observation layer's dynamics cause similar intractabilities.
\item We compare the proposed deep GP for IRL with existing approaches in benchmark tests as well as newly defined tests with more challenging rewards.
\end{itemize}

In the following, we review the problem of inverse reinforcement learning (IRL). The list of notations used throughout the paper is summarized in Table \ref{tab:notaions}.


\begin{table}[t!]
    \centering
    \caption{Summary of notations.}\label{tab:notaions}
\begin{tabularx}{\linewidth}{l @{\hspace{1.5em} \hspace{0.1em}}X}
  \toprule
  $t$ & time step $t\in \{1,2,\cdots,T\}$\\
  $h$ & index of demonstrations $h\in\{1,\cdots,H\}$\\
  $s,a$ & state $s\in\mathcal{S}$, action $a\in \mathcal{A}$\\
  $\gamma$& RL discount factor, $\gamma\in (0,1)$\\
  $\brwd$  & reward vector for all states $s\in \mathcal{S}$                  \\%
  $r(\cdot)$       & reward function for states $\mathcal{S}\mapsto \mathbb{R}$          \\%
  $Q(\cdot)$       & Q-value function $\mathcal{S}\times\mathcal{A}\mapsto\R$ \\
  $\mathcal{Q}$& variational distribution\\
  $V(\cdot)$       & state value function $\mathcal{S}\mapsto\R$ \\
  $\pi,\hat{\pi},\pi^*$ & policy $\mathcal{S}\mapsto\mathcal{A}$, and the corresponding estimated and optimal versions\\ 
  $\X$&  $m_0$-dimensional feature matrix for $|\mathcal{S}|=n$ discrete states, $ \X\in\R^{n,m_0}$\\
  $\x_i,\x^m$& features for $i$th state, $[\X]_{i,:}=\x_i$, the $m$th feature for all states, $[\X]_{:,m}=\x^m$\\
  $k_\bftheta$ & covariance function parametrized by $\btheta$\\
  $K_{\X,\X}$& covariance matrix, $[K_{\X,\X}]_{i,j}\!=\!k_\btheta(\x_i,\x_j)$ \\
  $\bB$& latent state matrix with $m_1$-dimensional features, $\bB\in\R^{n,m_1}$\\
  $\bD$& $\bB$ with Gaussian noise, $\bD\in\R^{n,m_1}$\\
  $\bW,\bZ$& inducing inputs, $\bW\!\in\!\R^{K,m_0}, \bZ\!\in\!\R^{K,m_1}$\\
  $\bV,\bfv$& inducing outputs,$\bV\in\R^{K,m_1}$,  $\bfv\in\R^{K}$\\ 
  tilde $\tilde{}$ & mean of the corresponding variable distribution, e.g., $\tilde{\bfv},\tilde{\bv},\tilde{\bD}$\\
  \bottomrule%
\end{tabularx}%
\end{table}

\subsection{Inverse Reinforcement Learning}
\label{sec:prelim}

The Markov Decision Process (MDP) is characterized by  $\{\mathcal{S},\mathcal{A}, \mathcal{T},\gamma,\brwd\}$, which represents the state space, action space, transition model, discount factor, and reward function, respectively. 

Take robot navigation as an example. The goal is to travel to the goal spot while avoiding stairwells. The state describes the current location and heading. The robot can choose actions from going forward or backward, turning left or right. The transition model specifies $p(s_{t+1}|s_t,a_t)$, i.e., the probability of reaching the next state given the current state and action, which accounts for the kinematic dynamics. The reward is +1 if it achieves the goal, -1 if it ends up in the stairwell, and 0 otherwise. The discount factor, $\gamma$, is a positive number less than 1, e.g., 0.9, to discount the future rewards. The optimal policy is then given by maximizing the expected reward, \ie
\begin{equation} \pi^*=\arg\min_{\pi}\E\left[\sum_{t=0}^\infty\gamma^tr(s_t)|\pi\right] .
\end{equation}

The IRL task is to find the reward function $r^*$ such that the induced optimal policy matches the demonstrations, given $\{\mathcal{S},\mathcal{A}, \mathcal{T},\gamma\}$ and $\bM=\{\zeta_1,...,\zeta_H\}$, where $\zeta_h=\{(s_{h,1},a_{h,1}),...,(s_{h,T},a_{h,T})\}$ is the demonstration trajectory, consisting of state-action pairs. Under the \textbf{linearity assumption}, the feature representation of states forms the linear basis of reward, $r(s)=\bw^\top \phi(s)$, where $\phi(s):\mathcal{S}\mapsto \R^{m_0}$ is the $m_0$-dimensional mapping from the state to the feature vector. From this definition, the \emph{expected reward} for policy $\pi$ is given by:
\begin{equation*}
\E\left[\sum_{t=0}^\infty\gamma^tr(s_t)|\pi\right]=\bw^\top\E\left[\sum_{t=0}^\infty\gamma^t\phi(s_t)|\pi\right]
\end{equation*}
where $\mu(\pi)=\E\left[\sum_{t=0}^\infty\gamma^t\phi(s_t)|\pi\right]$ is the feature expectation for policy $\pi$. The reward parameter $\bw^*$ is learned such that
\begin{equation}
\bw^{*\top}\mu(\pi^*)\geq \bw^{*\top}\mu(\pi), \forall\pi
\end{equation}
a prevalent idea that appears in the maximum margin planning (MMP) \citep{ratliff2006maximum} and feature expectation matching \citep{syed2007game}. 

Motivated by the perspective of expected reward that parametrizes the policy class, the maximum entropy (MaxEnt) model \citep{ziebart2008maximum} considers a stochastic decision model, where\emph{ the optimal policy randomly chooses the action} according to the associated reward: 
\begin{equation}
p(a|s)=\exp\{Q^*(s,a;\brwd)-V^*(s;\brwd)\}
\end{equation}
where $V(s;\brwd)=\log\sum_{a}\exp (Q(s,a;\brwd))$ follows the Bellman equation, $Q(s,a;\brwd)$ and $V(s;\brwd)$ are measures of how desirable are the corresponding state $s$ and state-action pair $(s,a)$ under rewards $\brwd$. In principle, for a given state $s$, the best action corresponds to the highest Q-value, which represents the ``desirability'' of the action. Assuming independence among state-action pairs from demonstrations, the likelihood of the demonstration corresponds to the joint probability of taking a sequence of actions $a_{i,t}$ under states $s_{h,t}$ according to the Bellman equation:
\begin{align}
&p(\bM|\brwd)=\prod_{h=1}^H\prod_{t=1}^T p(a_{h,t}|s_{h,t})\nonumber\\
&=\exp \left(\sum_{h=1}^H \sum_{t=1}^T \big(Q(s_{h,t},a_{h,t};\brwd)-V(s_{h,t};\brwd)\big)\right).
\label{equ:irl_maxent}
\end{align}
Though directly optimizing the above criteria with respect to $\brwd$ is possible, it does not lead to generalized solutions transferrable in a new test case where no demonstrations are available; hence, we need a ``model'' of $\brwd$. MaxEnt assumes linear structure for rewards, while GPIRL \citep{levine2011nonlinear} uses GPs to relate the states to rewards. In Section \ref{subsec:gp}, we will give a brief overview of GPs and GPIRL.

\section{The Model}
\label{sec:model}

In this section, we start by discussing the reward modeling through Gaussian processes (GPs) following \citep{levine2011nonlinear}, proceed to incorporate the representation learning module as additional GP layers, then develop our variational training framework and, finally, develop the transfer between tasks.

\subsection{Gaussian Process Reward Modeling}
\label{subsec:gp}
We consider the setup of \emph{discretizing the world} into $n$ states. 
Assume that observed state-action pairs (demonstrations) $\bM=\{\zeta_1,\dots,\zeta_h\}$ are \emph{generated} by a set of $m_0-$dimensional state features $\X \in \R^{n,m_0}$ through the reward function $r$. 
Throughout this paper we denote points (rows of a matrix) as $[\X]_{i,:}=\x_i$, features (columns) as $[\X]_{:,m}=\x^m$ and single elements as $[\X]_{i,m}=x_i^m$. 
%
%

In this modeling framework, the reward function $r$ plays the role of an unknown mapping, thus we wish to treat it as \emph{latent} and keep it flexible and non-linear. Therefore, we follow \citep{levine2011nonlinear} and model it with a zero-mean GP prior \citep{rasmussen2006gaussian}: 
\begin{equation*}
r \sim \mathcal{GP}(0,k_\bftheta(\x_i,\x_j)),
\end{equation*} 
where $k_\bftheta$ denotes the covariance function, 
e.g. $k_\bftheta(\x_i, \x_j) = \sigma_k^2 e^{-\frac{\xi}{2} (\x_i-\x_j)^\top (\x_i-\x_j)}$, 
$\bftheta=\{\sigma_k, \xi\}$. Given a finite amount of data, this induces the probability $\brwd|\X,\bftheta \sim \mathcal{N}(\bzero, K_{\X\X})$, where $\brwd \triangleq r(\X)$ and the covariance matrix is obtained by $[K_{\X\X}]_{i,j}=k_\bftheta(\x_i,\x_j)$. The GPIRL training objective comes from integrating out the latent reward: 
\begin{equation}
\label{eq:GPIRL_likelihood}
p(\bM|\X) = \int p(\bM | \brwd) p(\brwd | \X, \bftheta) \text{d}\brwd
\end{equation}
and maximizing over $\bftheta$, which we drop from our expressions from now on. 

The above integral is intractable, because $p(\bM|\brwd)$ has the complicated expression of \eqref{equ:irl_maxent} (this is in contrast to the tranditional GP regression where $\bM|\brwd$ is a Gaussian or other simple distribution). This can be alleviated using the approximation of \citep{levine2011nonlinear}. We will describe this approximation in the next section, as it is also used by our approach. Notice that all latent function instantiations are linked through a joint multivariate Gaussian. Thus, prediction of the function value $r^*=r(\x^*)$ at a test input $\x^*$ is found through the conditional 
\begin{equation*}
r^*|\brwd,\X,\x^*\sim \N\!\left(\!K_{\x^*\X}K_{\X\X}^{-1}\brwd, k_{\x^*\x^*} \!-\!K_{\x^*\X}K_{\X\X}^{-1}K_{\X\x^*}\!\right)
\end{equation*}

%

 As can be seen, the prediction $r(\mathbf{x}^*)$ is reliant on the \emph{effectiveness of feature representation}: states with features close in Euclidean distance are assumed to be associated with similar rewards. This motivates our novel deep GP-IRL method which is obtained by considering additional layers, as we will describe next.

\subsection{Incorporating the Representation Learning Layers}

The traditional model-based IRL approach is to learn the latent reward $\brwd$ (operating on fixed state features $\bX$) that best explains the demonstrations $\bM$. In this paper we wish to additionally and simultaneously uncover a highly descriptive state feature representation. To achieve this, we introduce a \emph{latent} state feature representation $\bB = \begin{bmatrix} \bb^1,...,\bb^{m_1} \end{bmatrix} \in\R^{n, m_1}$. $\bB$ constitutes the instantiations of an introduced function $b$ which is learned as a non-linear GP transformation from $\X$. To account for noise we further introduce $\bD$ as the noisy versions of $\bB$, \ie $d_i^m = b_i^m + \epsilon$, $\epsilon \sim \mathcal{N}(0, \lambda^{-1})$. 

 Importantly, rather than performing two steps of learning separately (for the GPs on $r$ and on $b$), we nest them into a single objective function, to maintain the flow of information during optimization. This results in a deep GP whose top layers perform representation learning and lower layers perform model-based IRL. Fig. \ref{fig:gpirl_overview} outlines our model, Deep Gaussian Process for Inverse Reinforcement Learning (\textbf{DGP-IRL}).  
By using $\x^m$ to represent the $m$-th column of $\bX$, and similarly for $\bD,\bB$ the full generative model is written as follows:
\begin{align}
& p(\bM,\brwd,\bD,\bB|\X)\nonumber\\
& = \underbrace{p(\bM|\brwd)}_{\text{IRL}}   \underbrace{p(\brwd|\bD)}_{\mathcal{GP}(\bzero, k^{r}(\bd_i,\bd_j))} \; \; \underbrace{p(\bD|\bB)}_{\text{Gaussian noise}} \; \; \underbrace{p(\bB|\bX)}_{\mathcal{GP}(\bzero, k^{b}(\x_i,\x_j))} \nonumber\\
&= e^{\sum_{h=1}^H \sum_{t=1}^T \big(Q(s_{h,t},a_{h,t};\brwd)-V(s_{h,t};\brwd)\big)}
\N(\brwd|\bzero,K_{\bD\bD})\nonumber\\
&\;\;\;\;\;\;\;\quad\prod_{m=1}^{m_1}\N(\bd^m|\bb^m,\lambda^{-1}\bI)\N(\bb^m|\bzero,K_{\bX\bX}),  \label{eq:dgpirl_joint}
\end{align}
where the IRL term $p(\bM|\brwd)$ takes the form of \eqref{equ:irl_maxent} as suggested by \cite{ziebart2008maximum}. $K_{\bX\bX}$ and $K_{\bD\bD}$ are the covariance matrices in each layer, constructed with covariance functions $k^{b}$ and $k^{r}$ respectively. Compared to GPIRL the proposed framework has substantial gain in flexibility by introducing the abstract representation of states in the hidden layers $\mathbf{B}, \mathbf{D}$. Note that the model in Fig. \ref{fig:gpirl_overview} can be extended in depth by introducing additional hidden layers and connecting them with additional GP mappings; it is only for illustration simplicity that we base our derivation on the two layered structure.  

We can compress the statistical power of each generative layer into a set of auxiliary variables within a \emph{sparse GP framework} \citep{snelson2005sparse}. Specifically, we introduce \emph{inducing outputs and inputs}, denoted by $\bfv \in \R^K$ and $\bZ \in \R^{K,m_1}$ respectively for the lower layer and by $\bV \in \R^{K,m_1}$ and $\bW\in \R^{K,m_0}$ for the top layer (as in Fig. \ref{fig:gpirl_overview}). The inducing outputs and inputs are related with the same GP prior appearing in each layer. For example, $\bfv|\bZ\sim \mathcal{N}(\bzero, K_{\bZ\bZ})$ with $K_{\bZ\bZ} = k^r(\bZ,\bZ)$. By relating the original and inducing variables through the conditional Gaussian distribution, the auxiliary variables are learned to be \emph{sufficient statistics} of the GP. The augmented model, shown in Fig. \ref{fig:gpirl_new}, has the following definition:
\begin{align}
&p(\bM,\brwd,\bfv,\bB,\bD,\bV|\bX,\bZ,\bW) \label{equ:fitc_r_f}\\
&= p(\bM|\brwd)p(\brwd|\bfv,\bD,\bZ)p(\bfv|\bZ)p(\bD|\bB) p(\bB|\bV,\!\bX,\!\bW) \nonumber\\
&=
e^{\sum_{h=1}^H \sum_{t=1}^T \big(Q(s_{h,t},a_{h,t};\brwd)-V(s_{h,t};\brwd)\big)}\cdot\nonumber\\
&\; \; \;\;\;\N(\brwd|K_{\bD\bZ}K_{\bZ\bZ}^{-1}\bfv,\bSigma_r)
\N(\bfv|\bzero,K_{\bZ\bZ})\cdot\nonumber\\
&\; \; \; \prod_{m=1}^{m_1}\N(\bd^m|\bb^m,\lambda^{-1}\bI)
\N(\bb^m|K_{\bX\bW}K_{\bW\bW}^{-1}\bv^m\!,\bSigma_B\!)\nonumber
\end{align}
where we adopt the Fully Independent Training Conditional (FITC) to preserve the exact variances in $\bSigma_B = diag\left(K_{\bX\bX}-K_{\bX\bW}K_{\bW\bW}^{-1}K_{\bW\bX}\right)$, and the Deterministic Training Conditional (DTC) \citep{quinonero2005unifying} in $\bSigma_r = \bzero$ as in GPIRL to facilitate the integration of $\brwd$ in the training objective (see next section).

In the following, we will omit the inducing inputs $\bW,\bZ$ in the conditions, with the convention to treat them as model parameters \citep{damianou2013deep,damianou2015deep,kandemir2015asymmetric}. 
By selecting $K \ll n$ the complexity reduces from $\mathcal{O}(n^3)$ to $\mathcal{O}(nK^2)$. While DGP-IRL resolves the case when the outputs have complex dependencies with the latent layers, the training of the model based on variational inference requires gradients for the parameters, as in backpropagation, whose convergence can be improved by leveraging advancements in deep learning.

Additionally, in \mbox{DGP-IRL}, the role of auxiliary variables goes further than just introducing scalability. Indeed, as we shall see next, the auxiliary variables play a distinct role in our model, by forming the base of a variational framework for Bayesian inference.





\subsection{Variational Inference}
\label{subsec:training}

We wish to optimize the model evidence for training:
\begin{align}
p(\bM|\bX) = &\int p(\bM,\bfv,\brwd,\bV,\bD,\bB|\bX)\text{d} (\bfv,\brwd,\bV\!,\bD,\bB)\nonumber
\end{align}

However, this quantity is intractable. Firstly because the latent variables $\bD$ appear nonlinearly in the inverse of covariance matrices. Secondly, because the latent rewards $\bfv,\brwd$ relate to the observation $\bM$ through the reinforcement learning layer; the choice of $\bSigma_r = \bzero$ in \eqref{equ:fitc_r_f} does not completely solve this problem because in DGP-IRL there is additional uncertainty propagated by the latent layers. This indicates that Laplace approximation is not practical, neither is the variational method employed for deep GP \citep{damianou2013deep,kandemir15asymmetric}, where the output is related to the latent variable in a simple regression framework. 

To this end, we show that we can derive an analytic lower bound on the model evidence by constructing a variational framework using the following special form of variational distribution:
\begin{align}
\mathcal{Q}&=q(\bfv)q(\bD)q(\bB)q(\bV), \; \text{with}:\label{equ:Q}\\
q(\bfv)&=\delta(\bfv-\tilde{\bfv})\nonumber\\
q(\bB)&=p(\bB|\bV,\bX) \label{equ:qB}\\
q(\bD)&=\prod_{m=1}^{m_1}\delta\left(\bd^m-K_{\bX\bW}K_{\bW\bW}^{-1}\tilde{\bv}^m\right)\\
q(\bV)&=\prod_{m=1}^{m_1}\N\left(\bv^m|\tilde{\bv}^m,\bG^m\right)
\end{align}
The delta distribution is equivalent to taking the mean of normal distributions for prediction, which is reasonable in the context of reinforcement learning \citep{levine2011nonlinear}. Also note that the delta distribution is \emph{applied only in the bottom layer and not repeatedly}; therefore, representation learning is indeed being manifested in the latent layers. 

In addition, $q(\bB)$ matches the exact conditional $p(\bB|\bV,\bX)$ so that these two terms cancel in the fraction of \eqref{log_lower:1} and the number of variational parameters is minimized, as in \citep{titsias2010bayesian}. As for $q(\bD)$, it is chosen as delta distributions such that combined with $\bSigma_r = \bzero$ the IRL term $p(\bM|\brwd)p(\brwd|\bfv,\bD)$ in \eqref{bayesrule} becomes tractable and information can flow through the latent layers $\bB,\bD$. 

The variational marginal $q(\bV)$ is factorized across its dimensions with fully parameterized normal densities, as in \citep{damianou2013deep}. Notice that $\tilde{\bfv}$ and $\tilde{\bv}$ are the mean of the inducing outputs (Table~\ref{tab:notaions}), corresponding to pseudo-inputs $\bZ$ and $\bW$, where $\bZ$ (initialized with random numbers from uniform distributions \citep{sutskever2013importance}) be learned to further maximize the marginaliked likelihood, and $\bW$ is chosen as a subset of $\bX$. 

The variational means of $\mathbf{D}$ can be augmented with input data, $\mathbf{X}$, to improve stability during training \citep{DuvRipAdaGha2014}. \\


\textbf{The variational lower bound}, $\Log$, follows from the Jensen's inequality, and can be derived analytically due to the choice of variational distribution $\mathcal{Q}$ (see Section 2 of the Appendix for details):
\begin{strip}
\begin{align}
&\log p(\bM|\bX)= \log \int \underbrace{\!\!\!\!\!\!\!\! p(\bM|\brwd)p(\brwd|\bfv,\bD)\!\!\!\!\!\!}_{\!\!\!\!\! \!\!\!\!\!\!p(\bM|\brwd=K_{\bD\bZ}K_{\bZ\bZ}^{-1}\bfv)\text{ by DTC: $\bSigma_r\!\!=\!\bzero$}\!\!\!\!\!\!\!\!\!\!}p(\bfv)p(\bD|\bB)p(\bB|\bV,\X)p(\bV)\text{d} (\brwd,\bfv,\bV,\bD,\bB)\label{bayesrule}\\
&\geq \int q(\bfv)q(\bD)p(\bB|\bV,\X)q(\bV)\log \frac{p(\bM|K_{\bD\bZ}K_{\bZ\bZ}^{-1}\bfv)p(\bfv)p(\bD|\bB)p(\bV)}{q(\bfv)q(\bD)q(\bV)} \text{  by Jensen's ineq.}\label{log_lower:1}\\
&=\Log_\bM+\Log_\mathcal{G}-\Log_{\text{KL}}+\Log_\mathcal{B}-\frac{nm_1}{2}\log (2\pi\lambda^{-1}) \label{eq:jensens}
\end{align}
where 
\begin{align}
\Log_\bM&=\log p(\bM|K_{\tilde{\bD}\bZ}K_{\bZ\bZ}^{-1}\tilde{\bfv}))=\sum_{h=1}^H \sum_{t=1}^T \bigg(\!Q(s_{h,t},\!a_{h,t};K_{\tilde{\bD}\bZ}\!K_{\bZ\bZ}^{-1}\!\tilde{\bfv})\!-\!V(s_{h,t};\!K_{\tilde{\bD}\bZ}\!K_{\bZ\bZ}^{-1}\tilde{\bfv})\!\!\bigg)\label{equ:L_M}\\
\Log_\mathcal{G}&=\log p(\bfv=\tilde{\bfv}|\bZ)=\log\N(\bfv=\tilde{\bfv}|\bzero,K_{\bZ\bZ})
\label{equ:L_G}\\
\Log_{KL}&=\text{KL}(q(\bV)||p(\bV|\bW))=\sum_{m=1}^{m_1} \text{KL}(\N(\bv^m|\tilde{\bv}^m,\bG^m)||\N(\bv^m|\bzero,K_{\bW\bW}))\nonumber\\
\Log_{\mathcal{B}}&=-\frac{\lambda}{2} \sum_{m=1}^{m_1} \text{Tr}(\bSigma_\bB+K_{\X\bW}K_{\bW\bW}^{-1}\bG^mK_{\bW\bW}^{-1}K_{\bW\X})
\end{align}
\end{strip}
where $|K_{\bW\bW}|$ is the determinant of $K_{\bW\bW}$. $\tilde{\bD}=\begin{bmatrix}       
\tilde{\bd}^1,...,\tilde{\bd}^{m_1}
\end{bmatrix}$, where $\tilde{\bd}^m=K_{\bX\bW}K_{\bW\bW}^{-1}\tilde{\bv}^m$. $\Log_\bM$ is the term associated with RL. $\Log_\mathcal{G}$ is the Gaussian prior on inducing outputs $\bfv$.  $\Log_{KL}$ denotes the Kullback -- Leibler (KL) divergence between the variational posterior $q(\bV)$ to the prior $p(\bV)$, acting as a regularization term. The lower bound $\Log$ can be optimized with gradient-based methods, which are computed by backpropagation. In addition, we can find the optimal fixed-point equations for the variational distribution parameters $\tilde{\bv}^m,\bG^m$ for $q(\bV)$ using variational calculus, in order to raise the variational lower bound $\Log$ further (refer to Section 3 of the supplement for this derivation).

Notice that the approximate marginalization of all hidden spaces, in \eqref{eq:jensens}, approximates a Bayesian training procedure, according to which model complexity is automatically balanced through the Bayesian Occam's razor principle. Optimizing the objective $\Log$ turns the variational distribution $\mathcal{Q}$ into an approximation to the true model posterior.

\subsection{Transfer to New Tasks}
\label{subsec:prediction}

The inducing points provide a succinct summary of the data, by the property of FITC \citep{quinonero2005unifying}, which means only the inducing points are necessary for prediction. Given a set of new states $\X^*$, DGP-IRL can infer the latent reward through the full Bayesian treatment:
\begin{align}
&p(\brwd^*|\bX^*,\bX)=\int \Big\lbrace p(\brwd^*|\bfv,\bD^*)q(\bfv)p(\bD^*|\bB^*)\nonumber\\
&\quad\quad\quad p(\bB^*|\bV,\bX^*)q(\bV)\Big\rbrace \text{d}(\bfv,\bB^*,\bD^*,\bV)
\end{align}
Given that the above integral is computationally intensive to evaluate, a practical alternative adopted in our implementation is to use point estimates for latent variables; hence, the rewards are given by:
\begin{equation}
\brwd^*=K_{\bD^*\bZ}K_{\bZ\bZ}^{-1}\tilde{\bfv}
\end{equation}
where $\bD^*=[\bd^{1}_*,...,\bd^{m_1}_*]$, with $\bd^{m}_* = K_{\bX^*\bW}K_{\bW\bW}^{-1}\tilde{\bv}^m$. The above formulae suggest that instead of making inference based on $\bX$ layer directly as in \cite{levine2011nonlinear}, DGP-IRL first estimates the latent representation of the states, $\bD^*$, then makes GP regression using the latent variables.

%

\section{Experiments}
\label{sec:exp}

For the experimental evaluation, we employ the \emph{expected value difference} (EVD) as a metric of optimality, given by:
\begin{equation}
\E\left[\sum_{t=0}^\infty\gamma^tr(s_t)|\pi^*\right]-\E\left[\sum_{t=0}^\infty\gamma^tr(s_t)|\hat{\pi}\right],
\end{equation}
which is the difference between the expected reward earned under the optimal policy, $\pi^*$, given by the true rewards, and the policy derived from the IRL rewards, $\hat{\pi}$. Our software implementation is included in the supplementary.

\subsection{Object World Benchmark}
\label{subsec:object}

\begin{figure}[t!]
      \subfigure[Ground Truth]{%
    \includegraphics[width=0.312\columnwidth,trim=0mm 0mm 0mm 0mm,clip]{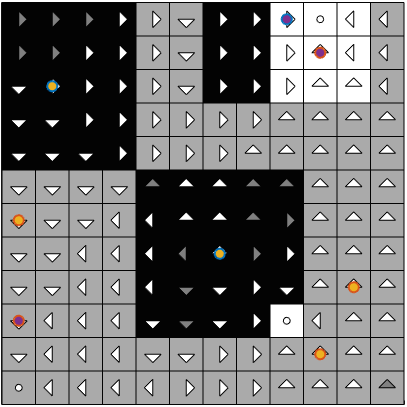} 
    }
    \subfigure[DGP-IRL]{%
        \includegraphics[width=0.312\columnwidth]{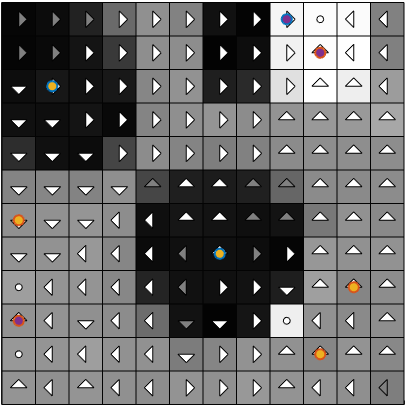}    
    }
    \subfigure[GPIRL]{%
  \includegraphics[width=0.312\columnwidth,trim=0mm 0mm 0mm 0mm,clip]{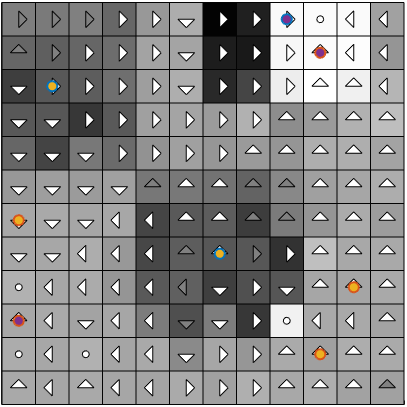} 
    }
    
    \subfigure[MWAL]{%
    \includegraphics[width=0.312\columnwidth,trim=0mm 0mm 0mm 0mm,clip]{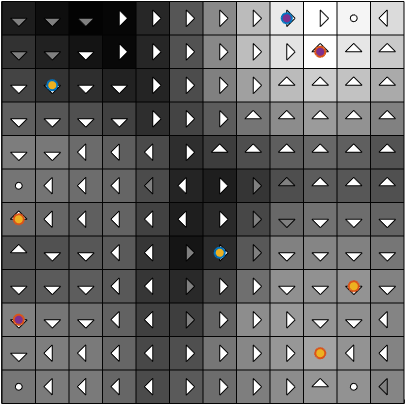} 
    }
    \subfigure[MaxEnt]{%
        \includegraphics[width=0.312\columnwidth]{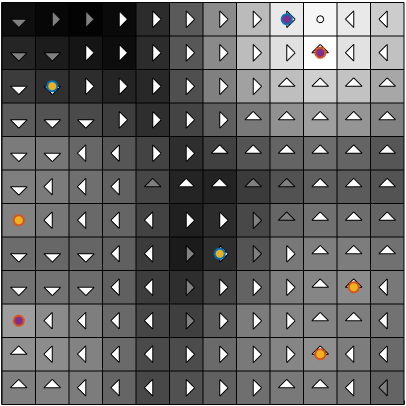}    
    }
    \subfigure[MMP]{%
  \includegraphics[width=0.312\columnwidth,trim=0mm 0mm 0mm 0mm,clip]{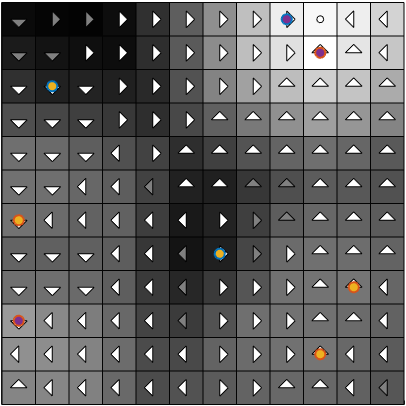} 
    }
     \caption{OW benchmark for IRL, evaluated for (a) DGP-IRL, (b) GPIRL, (c) MWAL, (d) MaxEnt, and (e) MMP, with 64 demonstrations and continuous features. Except for DGP-IRL, all the other algorithms are evaluated with the toolbox by \citet{levine2011nonlinear}. }\vspace{-.1in}
     \label{fig:obj_world}
   \end{figure}

The Object World (OW) benchmark, originally implemented by \cite{levine2011nonlinear}, is a $N\times N$ gridworld where dots of \emph{primary colors}, e.g., red and blue, and \emph{secondary colors}, e.g., purple and green, are placed on the grid at random, as shown in Fig. \ref{fig:obj_world}. Each state, i.e., grid block, is described by the shortest distances to dots among each color group. The latent reward is assigned such that if a block is 1 step within a red dot and 3 steps within a blue dot, the reward is +1; if it is 3 steps within a blue dot only, the reward is -1, and the reward is 0 otherwise. The agent maximizes its expected discounted reward by following a policy which provides the probabilities of actions (moving up/down/left/right, or stay still) at each state, subject to a transition probability.

The objective of the experiment is to compare the performances of DGP-IRL with previous methods as the number of demonstrations varies. Candidates that are evaluated include the Multiplicative Weights for Apprenticeship Learning (MWAL) \citep{syed2007game}, MaxEnt, MMP, which assume a linear reward function, and GPIRL, which is the state-of-the-art method on the benchmark. Linear models, as is shown in Fig. \ref{fig:obj_world}, cannot capture the complex structure, while GPIRL learns more accurate yet still noisy rewards, as limited by feature discriminability; DGP-IRL, on the contrary, makes inference closest to the ground truth even with limited data, thanks to the increased representational power and robust training through variational inference.

 \begin{figure}[h!]
 \vspace{-.15in}
      \subfigure[Training]{%
    \includegraphics[width=0.45\columnwidth,trim=0mm 0mm 0mm 0mm,clip]{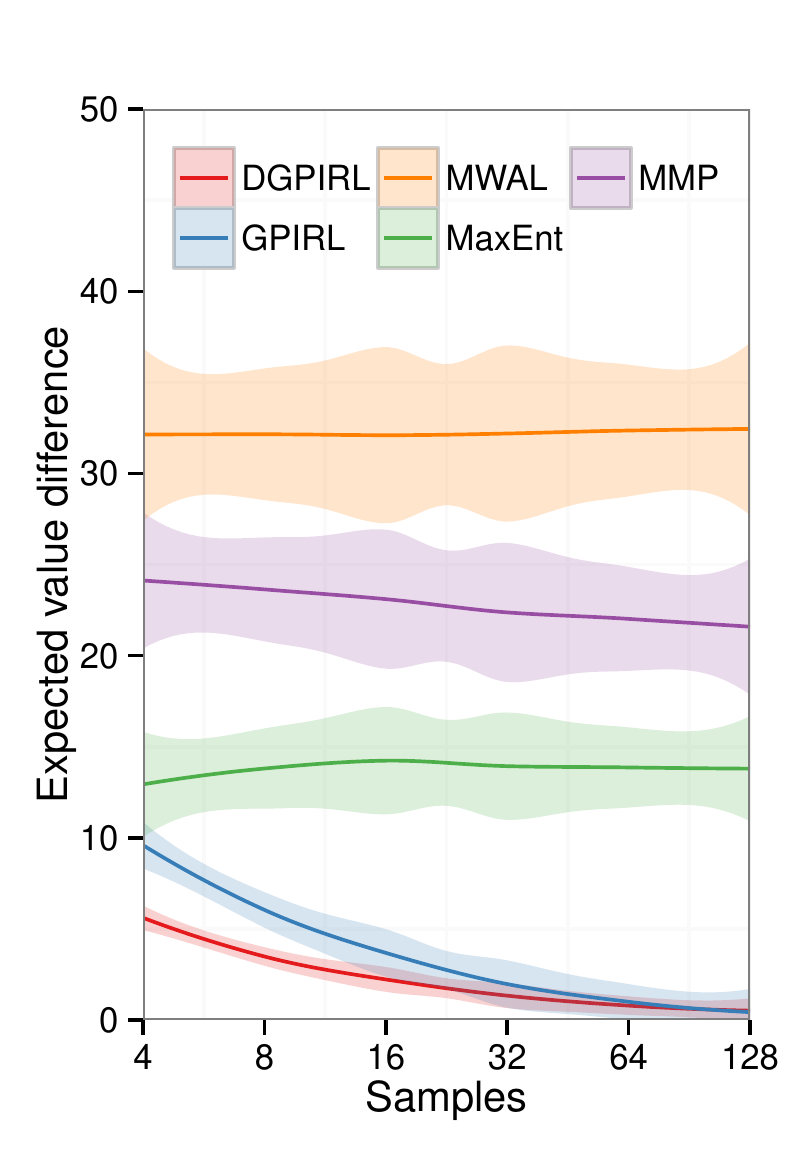} 
    }
    \subfigure[Transfer]{%
        \includegraphics[width=0.45\columnwidth]{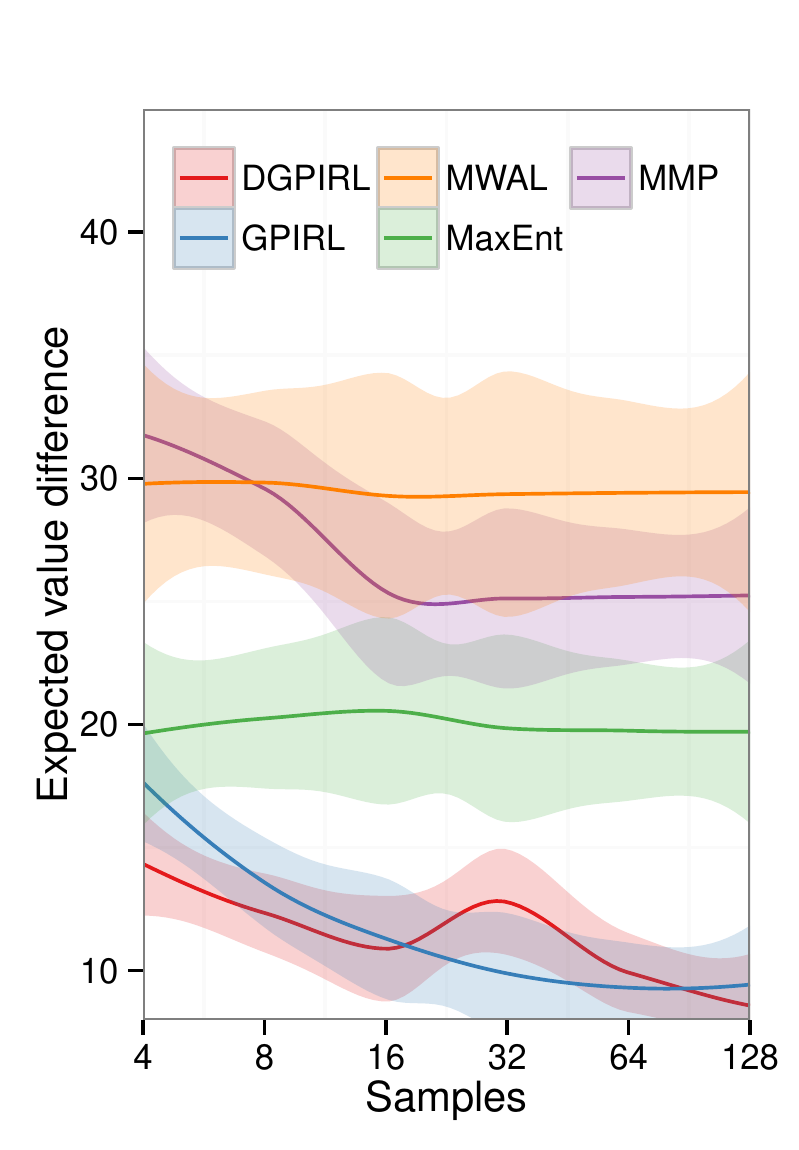}    
    }
     \caption{Plots of EVD in the training (a) and transfer (b) tests for the OW benchmark, where continuous features are employed. }
     \label{fig:obj_world_evd}
   \end{figure}

\begin{figure}[t]
      \subfigure[Training EVD]{%
    \includegraphics[width=0.45\columnwidth,trim=0mm 0mm 0mm 0mm,clip]{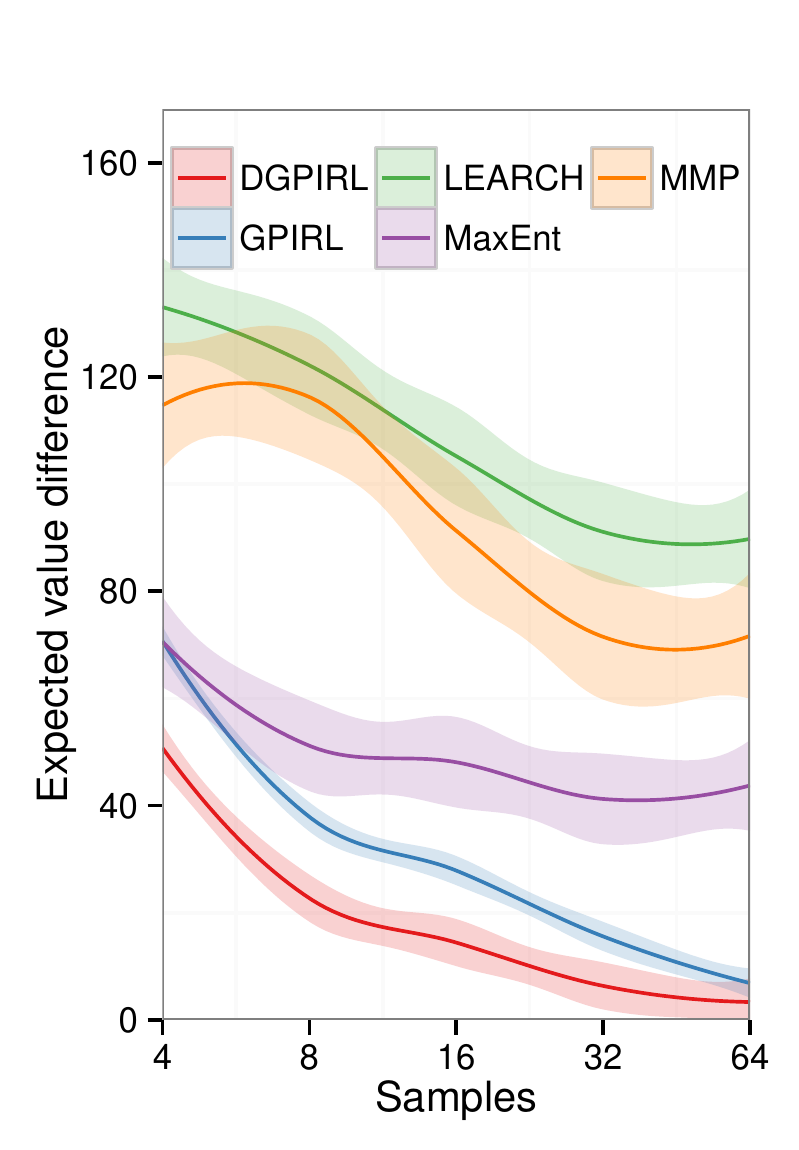} 
    \label{fig:highway_evd}
    }
    \subfigure[Speeding Prob.]{%
        \includegraphics[width=0.45\columnwidth]{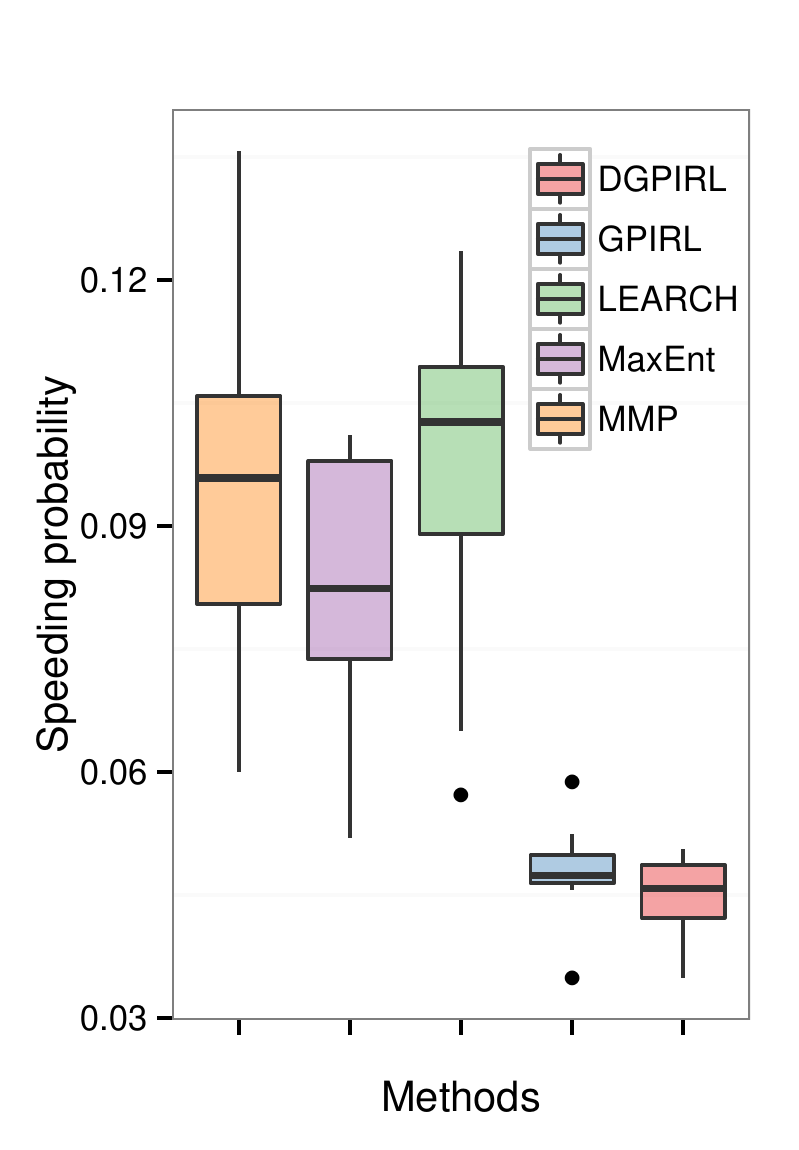}    
        \label{fig:highway_speed}
    }
     \caption{Plots of EVD in the training (a) and the probability of speeding (with 64 demonstrations) (b) in the highway driving simulation benchmark, with three lanes and 32 car lengths.}
     \label{fig:highway_evd_speed}
   \end{figure}

Additionally, the transferability test is carried out by examining EVD in a new world where no demonstrations are available, which requires the ability of knowledge transfer from the previous learning scenario. DGP-IRL outperforms GPIRL and other models in both the training and transfer cases, and the improvement is obvious as more data is accessible (Fig. \ref{fig:obj_world_evd}). The shaded area (Fig. \ref{fig:obj_world_evd},  \ref{fig:bin_world_evd},  \ref{fig:highway_evd}) are the standard deviation of EVD among independent experiments, which reflect that DGP-IRL and GPIRL are more reliable.

\subsection{Binary World Benchmark}
\label{subsec:binary}

Though the rewards in OW are nonlinear functions of the features, they form separated clusters in the subspace spanned by two dimensions, i.e., distances to the nearest red and blue dots. Binary world (BW) is a benchmark introduced by \cite{wulfmeier2015deep} whose reward depends on combinatorics of features. More specifically, in a world of $N\times N$ plane where each block is randomly assigned with either a blue or red dot, the state is associated with the +1 reward if there are 4 blues in the $3\times 3$ \emph{neighborhood}, -1 for 5 blues, and 0 otherwise. The feature represents the color of the 9 dots in the neighborhood. BW sets up a challenging scenario, where states that are maximally separated in feature space can have the same rewards, yet those that are close in euclidean distance may have opposite rewards.

\begin{figure}[t!]
      \subfigure[Ground Truth]{%
    \includegraphics[width=0.312\columnwidth,trim=0mm 0mm 0mm 0mm,clip]{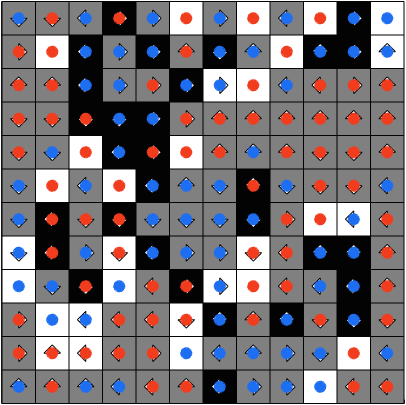} 
    }
    \subfigure[DGP-IRL]{%
        \includegraphics[width=0.312\columnwidth]{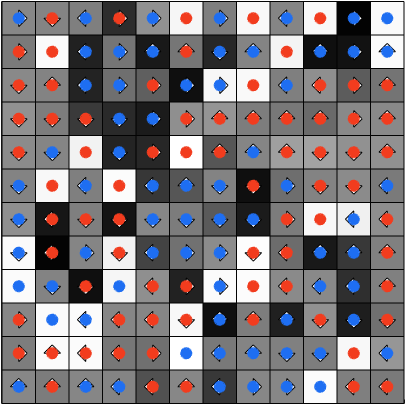}    
    }
    \subfigure[GPIRL]{%
  \includegraphics[width=0.312\columnwidth,trim=0mm 0mm 0mm 0mm,clip]{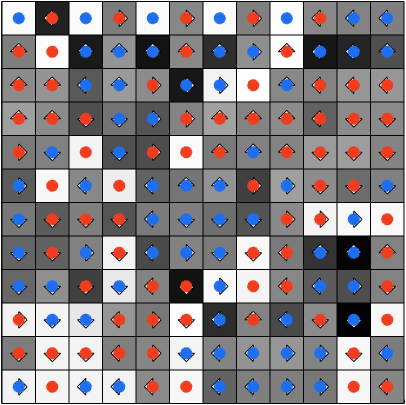} 
    }
    
    \subfigure[LEARCH]{%
    \includegraphics[width=0.312\columnwidth,trim=0mm 0mm 0mm 0mm,clip]{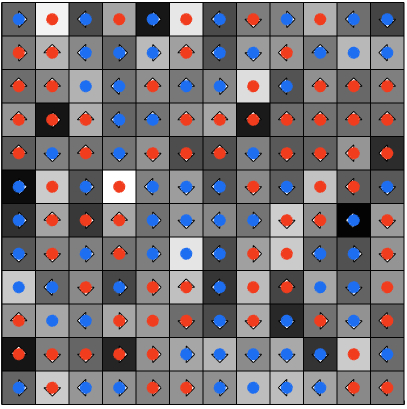} 
    }
    \subfigure[MaxEnt]{%
        \includegraphics[width=0.312\columnwidth]{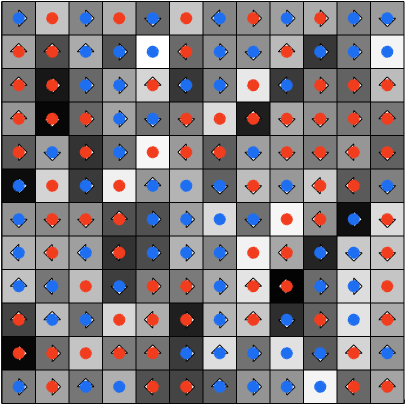}    
    }
    \subfigure[MMP]{%
  \includegraphics[width=0.312\columnwidth,trim=0mm 0mm 0mm 0mm,clip]{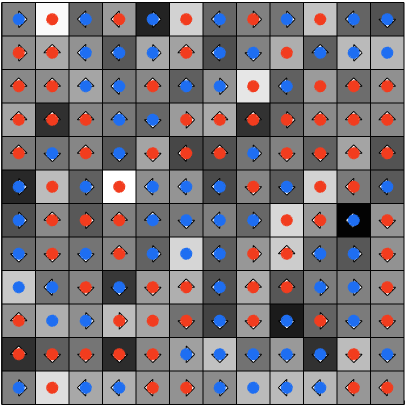} 
    }
     \caption{BW benchmark evaluated with 128 demonstrated traces for DGP-IRL, GPIRL, LEARCH \cite{ratliff2009learning}, MaxEnt, and MMP. }   \vspace{.06in}
     \label{fig:bin_world}
   \end{figure}

The task is to learn the latent rewards of states given limited demonstrations, as is shown in Fig. \ref{fig:bin_world}. While linear models are limited by their capacity of representation, the results of GPIRL also deviate from the latent rewards as it cannot generalize from training data with the convoluted features. DGP-IRL, nevertheless, is able to recover the ground truth with the highest fidelity.

 \begin{figure}[t!]
      \subfigure[Input space $\bX$]{%
    \includegraphics[width=0.48\columnwidth,trim=0mm 0mm 0mm 0mm,clip]{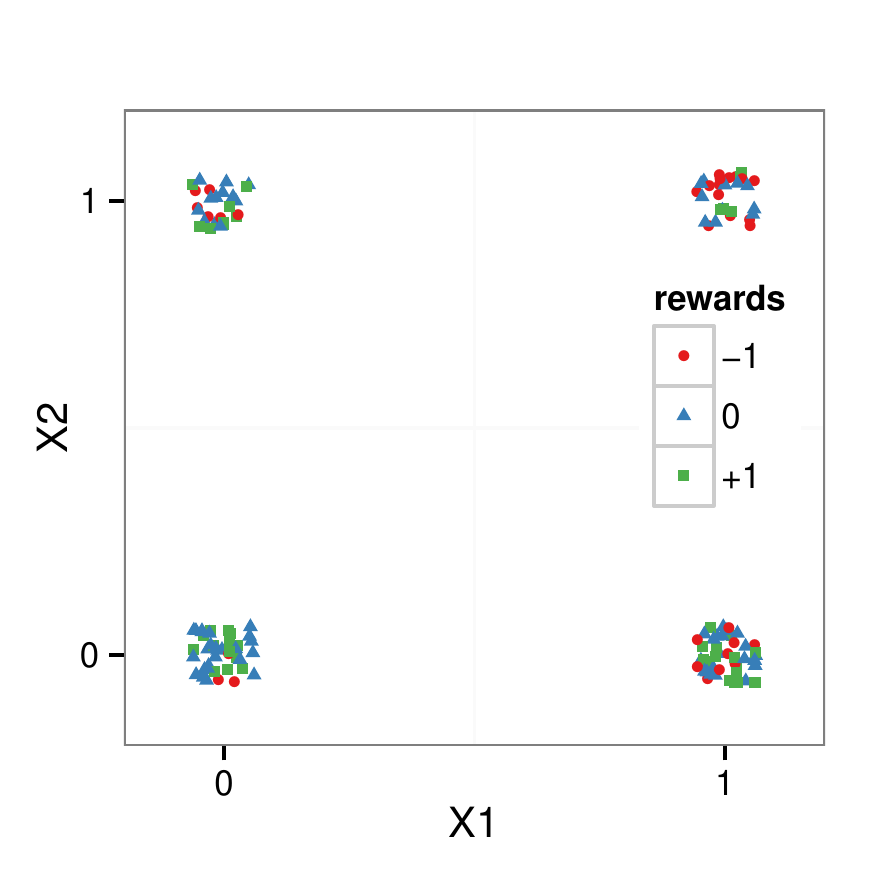} 
    }
    \subfigure[Latent space $\bD$]{%
        \includegraphics[width=0.48\columnwidth]{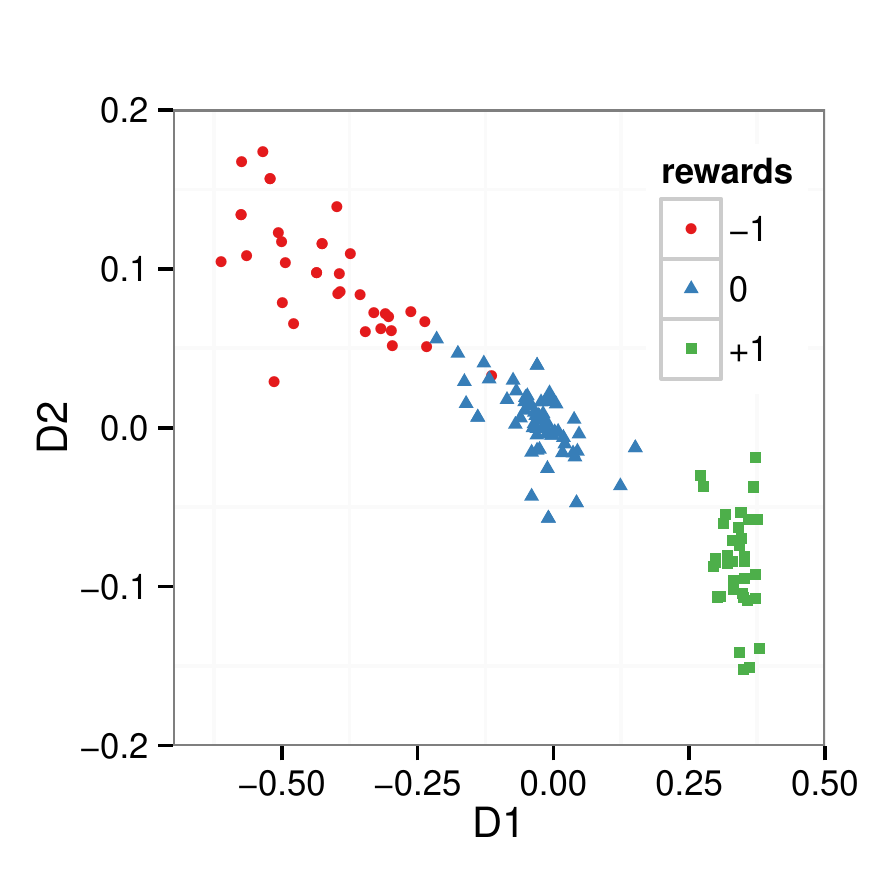}    
    }
     \caption{Visualization of points along two arbitrary dimensions in the (a) input space $\bX$ and (b) latent space $\bD$ of DGP-IRL. The points represent features of states and are color-coded with the associated rewards. The rewarads are entangled in the input space $\bX$ but separated in the latent space $\bD$.}
     \label{fig:bw_dx}
   \end{figure}
   
 \begin{figure}[t]
      \subfigure[Training]{%
    \includegraphics[width=0.45\columnwidth,trim=0mm 0mm 0mm 0mm,clip]{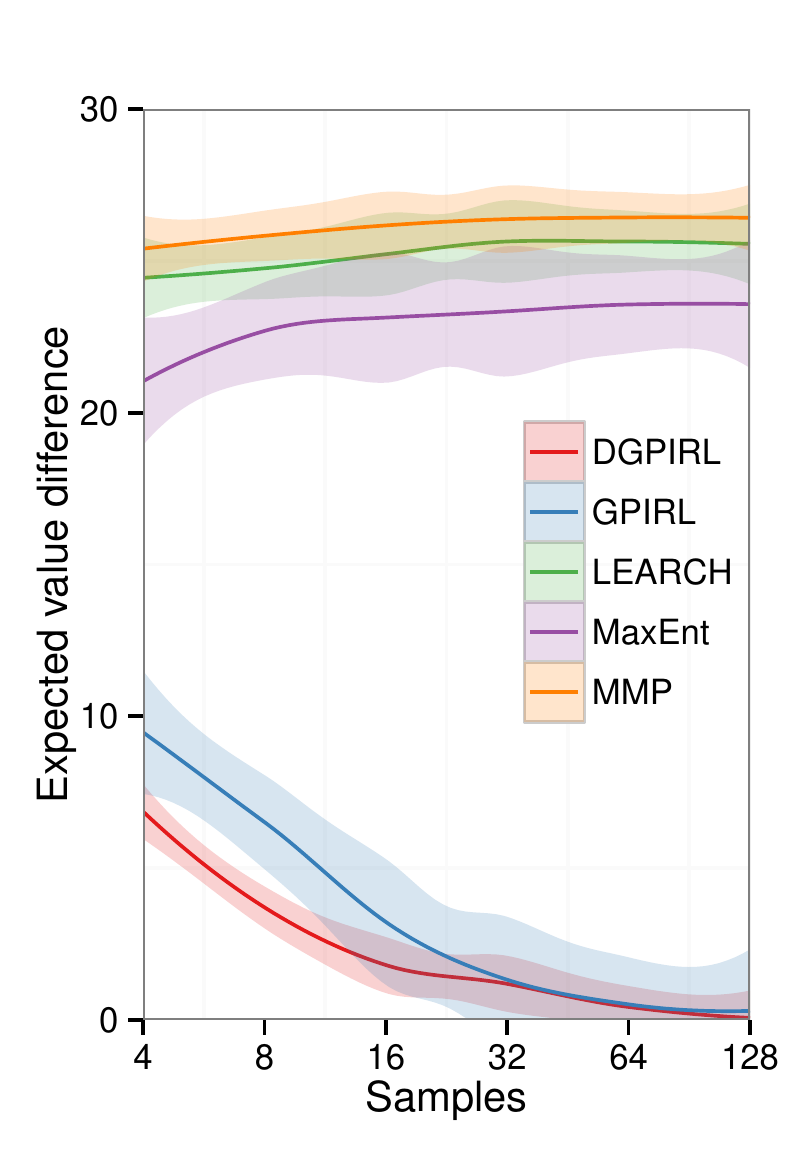} 
    }
    \subfigure[Transfer]{%
        \includegraphics[width=0.45\columnwidth]{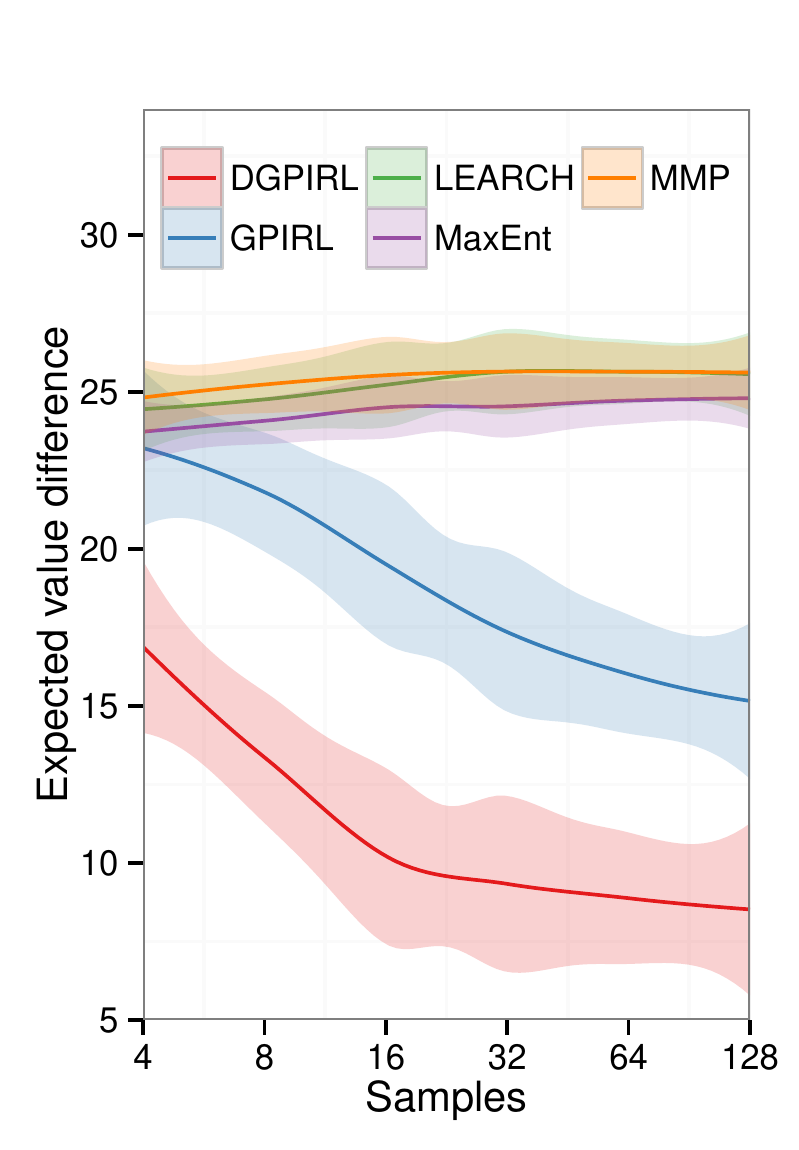}    
    }
     \caption{Plots of EVD in the training (a) and transfer (b) tests for the BW benchmark as the number of training samples varies. }
     \label{fig:bin_world_evd}
   \end{figure}

By successively warping the original feature space through the latent layers, DGP-IRL can learn an abstract representation that reveals the reward structure. As is illustrated in Fig. \ref{fig:bw_dx}, though the points are mixed up in the input space, making it impossible to separate those with the same rewards, their positions in the latent space clearly form clusters, which indicates that DGP-IRL has remarkably uncovered the mechanism of reward generation by simply observing the traces of actions.

The advantage of simultaneous representation and inverse reinforcement learning is demonstrated in Fig. \ref{fig:bin_world_evd}, which plots the EVD for the training and transfer cases. As the features are interlinked not only
with the reward but also with themselves in a very nonlinear way, this scenario is particularly challenging for linear models, such as LEARCH \citep{ratliff2009learning}, MaxEnt, and MMP. While both GPIRL and DGP-IRL have satisfactory performance in the training case, DGP-IRL significantly outperforms GPIRL in the transfer task, which indicates that the learned latent space is transferrable across different scenarios.

\subsection{Highway Driving Behaviors}
\label{subsec:highway}

Highway driving behavior modeling, as investigated by \cite{,levine2010feature, levine2011nonlinear}, is a concrete example to examine the capacity of IRL algorithms in learning the underlying motives from demonstrations based on a simple simulator. In a three-lane highway, vehicles of specific class (civilian or police) and category (car or motorcycle)  are positioned at random, driving at the same constant speed. The robot car can switch lanes and navigate at up to three times the traffic speed. The state is described by a continuous feature which consists of the closest distances to vehicles of each class and category in the same lane, together with the left, right, and any lane, both in the front and back of the robot car, in addition to the current speed and position. 

The goal is to navigate the robot car as fast as possible, but avoid speeding by checking that the speed is no greater than twice the current traffic when the police car is 2 car lengths nearby. As the reward is a nonlinear function determined by the current speed and distance to the police, linear models are outrun by GPIRL and DGP-IRL. Performance generally improves with more demonstrations, and DGP-IRL remains to yield the policy closest to the optimal in EVD, and with minimal probability of speeding, as illustrated in Fig. \ref{fig:highway_evd} and \ref{fig:highway_speed}, respectively.

\section{Conclusion and Future Work}
\label{sec:conclusion}

DGP-IRL is proposed as a solution for IRL based on deep GPs. By extending the structure with additional GP layers to learn the abstract representation of the state features, DGP-IRL has more representational capability than linear-based and GP-based methods. Meanwhile, the Bayesian training approach through variational inference guards against overfitting, bringing the advantage of automatic capacity control and principled uncertainty handling. 

Our proposed DGP-IRL outperforms state-of-the-art methods in our experiments, and is shown to learn efficiently even from few demonstrations.
For future work, the unique properties of DGP-IRL enable easy incorporation of side knowledge (through priors on the latent space) to IRL, but our work also opens up the way for combining deep GPs with other complicated inference engines, \eg selective attention models \citep{gregor2015draw}. We plan to investigate these ideas in the future. Another promising future direction is to construct transferable models where the latent layer is relied on for knowledge sharing. 
Finally, we plan to investigate some of the many applications where DGP-IRL can prove especially beneficial, such as intelligent building and grid controls \citep{jin2017mod,jin2017virtual} and human-in-the-loop gamification \citep{ratliff2014social}.

\bibliography{dgpirl}
\bibliographystyle{plainnat}

\end{document}